# Sub- Dividing Genetic Method for Optimization Problems

**MasoumehVali**

Department of Mathematics, Dolatabad Branch, Islamic Azad University, Isfahan, Iran
E-mail: vali.masoumeh@gmail.com

**Abstract**

Nowadays, optimization problem have more application in all major but they have problem in computation. Computation global point in continuous functions have high calculation and this became clearer in large space .In this paper, we proposed Sub- Dividing Genetic Method(SGM) that have less computation than other method for achieving global points . This method userotation mutation and crossover based sub-division method that sub diving method is used for minimize search space and rotation mutation with crossover is used for finding global optimal points. In experimental, SGM algorithm is implemented on De Jong function. The numerical examples show that SGM is performed more optimal than other methods such as Grefensstette, Random Value, and PNG.

1.  Introduction

Optimization problems have several applications such as computer sciences, operations research, economics, and engineering design and control but most of them have problem in computational and are searching simple method for earning global points.

In the computer science field of artificial intelligence, a genetic algorithm (GA) is a search heuristic that mimics the process of natural evolution. This heuristic (also sometimes called a met heuristic) is routinely used to generate useful solutions to optimization and search problems. Genetic algorithms belong to the larger class of evolutionary algorithms (EA), which generate solutions to optimization problems using techniques inspired by natural evolution, such as inheritance, mutation, selection, and crossover. Genetic algorithms find application in bioinformatics, phylogenetic, computational science, engineering, economics, chemistry, manufacturing, mathematics, physics, pharmacy metrics and other fields. In optimization problem can combined with GA for finding optimal point in search space.

In this paper, we propose SGM to solve this problem in Simple-bounded Continuous Optimization Problem by simple algebra, without derivation. The SG Minimize Search space by subdividing and labeling method and when we can recognize limited of solution, we used GA by rotational mutation and crossover for finding final Global optimal point.

This paper starts with the description of related work in section 2. Section 3 gives the outline of Model and Problem Definition SGM. In section 4, we have text problems of SGM. In section 5, we present a schema for SGM. In section 6, we present schema analysis for SGM. Evaluation and conclusion is in sections 7 and 8.

2. **Related Work**

The genetic algorithm (GA) is a search heuristics that is routinely used to generate useful solutions to optimization and search problems and more and more methods based on GA has been applied to find optimal point in continuous functions in recent years. For example, Pratibha Bajpal and Manojkumar [1] have proposed an approach to solve Global Optimization Problems by GA on both continuous and discrete optimization problems and obtained. Devis Karaboga and Selcuk proposed A Simple and Global Optimization Algorithm for Engineering Problems: Differential Evolution Algorithm [2].
In 2009, Zhang et al. [6] introduced triangulation theory into GA by the virtue of the concept of relative coordinates genetic coding, designs corresponding crossover and mutation operator and Hayes and Gedeon [3] considered infinite population model for GA where the generation of the algorithm corresponds to a generation of a map. In 2011, Zhang et al. [4] introduced triangulation theory into GA by the virtue of the concept of relative coordinates genetic coding, designs corresponding crossover and mutation operator. In 2012, Meera Kapoor and Vaishali Wadhwa proposed Optimization of DE Jong's Function Using Genetic Algorithm Approach. Their purpose of this master thesis is to optimize/maximize de Jong's function1 in GA using different selection schemes (like roulette wheel, random selection, fit/elitist fit rank selection, tournament selection) and so on.[ 5]

3. **Model and Problem Definition of SGM**

In this section, at first, Model of SGM is expressed and implemented on two problems. All operations are traced step by step.

At present, Suppose F($x_1, x_2, x_3 \ldots, x_n$) with constraint $a_i \leq x_i \leq b_i$ for i=1,2,…,n. We want to earn global point in order to follow serial algorithm SGM:

**Step1:** Draw the diagrams for $x_i = b_i$ and $x_i = a_i$ for i=1, 2… n then we find crossing points which equals $n^2$ and h= min$\frac{|x_i| + |x_j|}{2}$ for $1 \leq i, j \leq n$.

**Step2:** Suppose that the point $(a_1, b_1, b_2 \ldots b_{n-1})$ is one of the crossing points. Consider the values of $\pm h$ gained by step1 and then do the algebra operations on this crossing point that are shown in Table 1. The maximum number of them in dimensional space is $2 * (\binom{n}{1} + \binom{n}{2} + \binom{n}{3} + \cdots + \binom{n}{n})$.

**Step 3:** The function value is calculated for all of points of step 2 and the value of these functions is compared with $f(a_1, b_1, b_2 \ldots b_{n-1})$. At the end, we will select the point which has the minimum value and we will call it $(c_1, c_2, c_3 \ldots c_n)$.

**Note:** If we want to find the optimal global max, we should select the maximum value.
**Step 4:** In this step, the equation 1 is calculated:

$$(c_1, c_2, c_3 \ldots c_n) - (a_1, b_1, b_2 \ldots b_{n-1}) = (d_1, d_2, d_3 \ldots d_n) \qquad (1)$$

**Step 5:** According to the result of step 4, the point $(a_1, b_1, b_2 \ldots b_{n-1})$ is labeled according to equation 2.

**Step 6:** Go to step 7 if all the crossing points were labeled, otherwise repeat the steps 2 to 5.

**Step 7:** in this step, complete labeling polytope is focused. In fact, apolytope will be chosen that have complete labeling in different dimensional that is shown in table 2.

**Step 8:** In this step, all sides of the selected polytope (from step 7) are divided into 2 according to equation 3 and we repeat steps 3 to 7 for new crossing points.

$$h = \min\left\{\frac{|x_i| + |x_j|}{2}\right\} \quad 1 \leq i, j \leq n \qquad (3)$$

**Step 9:** Steps 2 to 8 are repeated as $k + 1$ according to constraint 4 and result is a completed label polytope.

$$-10k \leq x_1, x_2, \ldots, x_n \leq 10k \quad k \in \mathbb{N} \qquad (4)$$

**Step 10:** Regarding the selected polytope from step 9, we get vertex of this polytope which has best value and call it S (If goal is global min, the best value is the point that has the lowest value. If goal is max; the best value is the point of the highest value)

**Step 11:** In this step, put $s' = f(s)$, $\vec{r_0} = (1,1,\ldots,1), (1,-1,1,\ldots,1),\ldots,$ or $(-1,-1,\ldots,-1)$ and $r_n = \alpha \vec{r_0}, = 0.1n$, n=1,2,…,10

**Step 12:** At present, move from point 'S' in direction of $\vec{r_0}$ vector with length of α. (notice: the direction of $\vec{r_0}$ vector must be inside the search space, also the α measurement depends on problem precision) and the endpoint of vector $r_n$ is called p.

**Step 13:** if f (p) better than S', then S=p and go to step 16 else go to step 14.

**Step 14:** Put $\beta = 0.1, 0.25, ...$, $e_n = (1,1,...,1), (1,-1,1,...,1), ...,$ and $(-1,-1,...,-1)$, $P_n = \beta e_n$.

**Step 15:** If f (p) is better than $S'$, then S=p else go to step 16.

**Step16:** If $r_n$ is in search space, go to step 17 else go to stop.

**Step17:** In this section, sides of polytope that have common point with $r_n$ are subdivided in halves that cause new points to be generated.

**Step 18:** Among the obtained points from step 17 and point 'S' select point which has best fitness; then, assign $s'$ =best value and s=point of best value then go to step 11 and repeat algorithms.

### 4. Implementing Test Problems by SGM

**Test problem1:**

Test problem: The minimization problem is formulated as follows:

$$\min_{x \in \mathbb{R}^2} f(x) = x_1^2 + x_2^2 - 18 \cos x_1 - 18 \cos x_2 \tag{5}$$

Its global minimum is equal to -2 and the minimum point is at (0, 0). There is approximately 50 local minima in the region bounded by the two constraints.

|  | Table 1: Initial Population of $f_1$ |
|---|---|
| Figure 1: Initial Population of $f_1$ | <table><tr><td>$h_1=2$, P(0):</td><td>$h_2=1$</td><td>P(1):</td><td></td><td>Solution</td></tr><tr><td>(-1,1)</td><td rowspan="4">$\xrightarrow{M}$</td><td>(0,0)</td><td>2</td><td rowspan="4">(0,0)</td></tr><tr><td>(1,1)</td><td>(0,0)</td><td>2</td></tr><tr><td>(-1,-1)</td><td>(0,0)</td><td>0</td></tr><tr><td>(1,-1)</td><td>(0,0)</td><td>1</td></tr></table> |

Figure 1: Initial Population of $f_1$

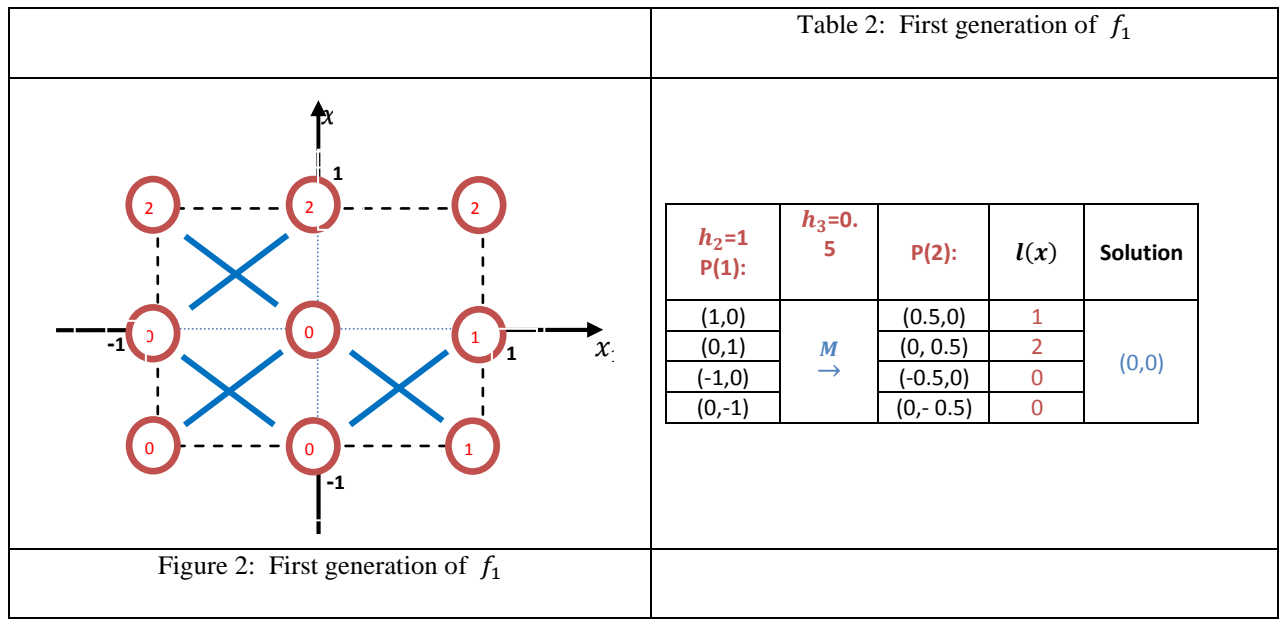

| | Table 2: First generation of $f_1$ |
|---|---|
| Figure 2: First generation of $f_1$ | |

| $h_2=1$ P(1): | $h_3=0.5$ | P(2): | $l(x)$ | Solution |
|---|---|---|---|---|
| (1,0) | | (0.5,0) | 1 | |
| (0,1) | $M \rightarrow$ | (0, 0.5) | 2 | (0,0) |
| (-1,0) | | (-0.5,0) | 0 | |
| (0,-1) | | (0,- 0.5) | 0 | |

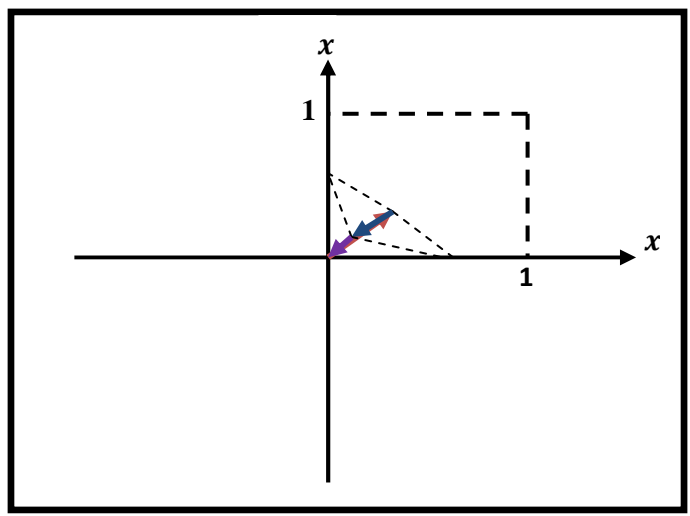

Figure3 : The process of rotational mutation and crossover on search space

Table 3: comparison between test problem #1 and other three

| Algorithms | Iteration | Optimal point | Best Solution | Standard deviation |
|---|---|---|---|---|
| SGM | 30 | (0,0) | | (0,0) |
| RS | 1000 | (0,0) | | (0,0) |
| **RSW**($x^{initial} =$ (14.0356,14.0356 )) | 500 | ( 0, 0. 004999 ) | (0,0) | ( 0, 0. 004999 ) |
| SA | 150 | (0,0) | | (0,0) |

**Test problem2:**

Consider Beale Function as follow:

$$f(x) = (1.5 - x_1 + x_1 x_2)^2 + (2.26 - x_1 + x_1 x_2^2)^2 + (2.625 - x_1 + x_1 x_2^3)^2, \quad (6)$$

$$-4.5 \leq x_i \leq 4.5, i = 1, 2.$$

The global minimum: $x^* = (3, 0.5)$, $f(x^*) = 0$. Beal function's graph is as follow.

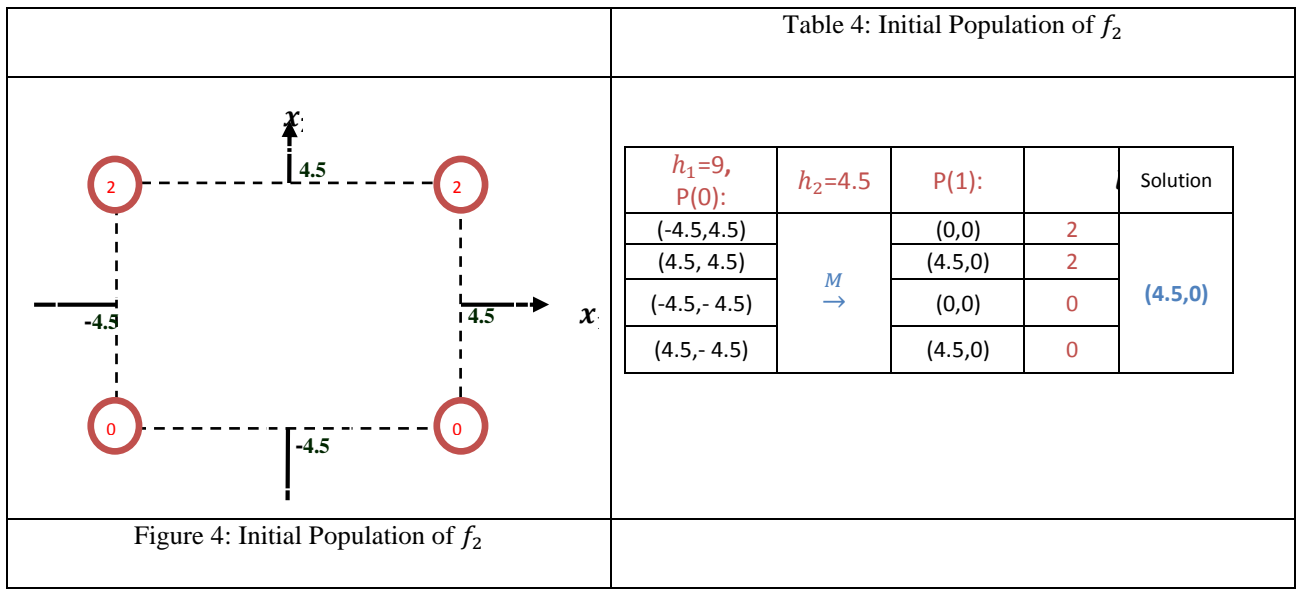

Figure 4: Initial Population of $f_2$

Table 4: Initial Population of $f_2$

| $h_1=9$, P(0): | $h_2=4.5$ | P(1): |   | Solution |
|---|---|---|---|---|
| (-4.5, 4.5) |  | (0,0) | 2 |  |
| (4.5, 4.5) | $\xrightarrow{M}$ | (4.5,0) | 2 | (4.5,0) |
| (-4.5,- 4.5) |  | (0,0) | 0 |  |
| (4.5,- 4.5) |  | (4.5,0) | 0 |  |

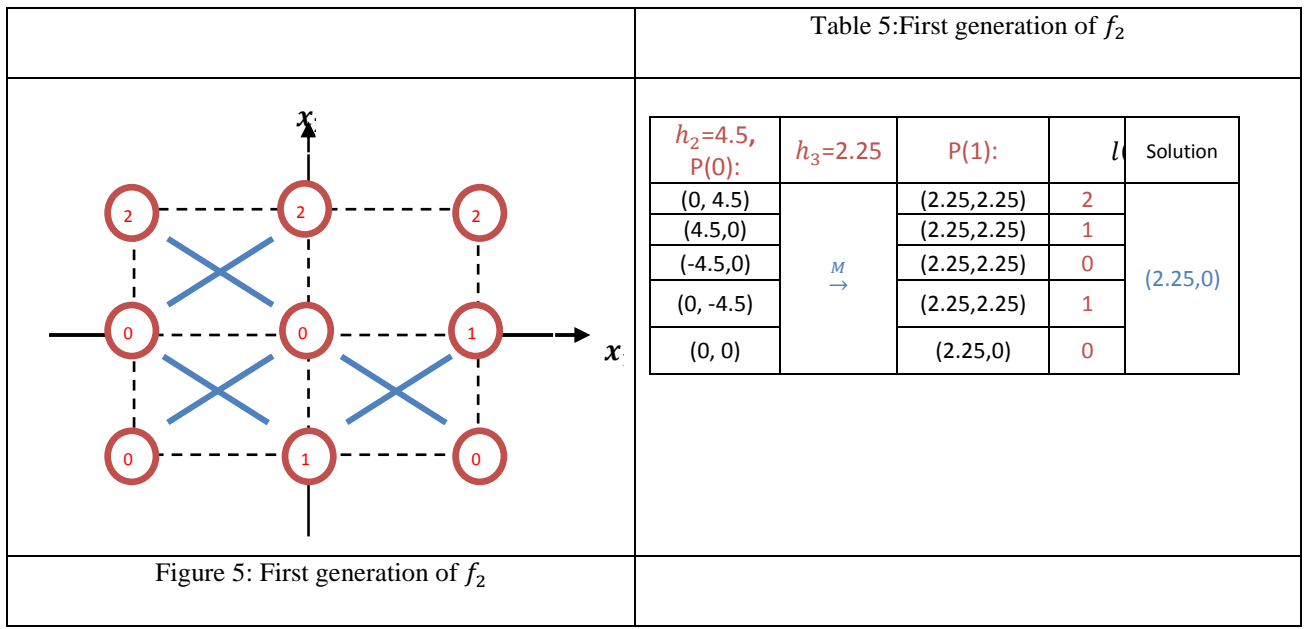

Figure 5: First generation of $f_2$

Table 5: First generation of $f_2$

| $h_2=4.5$, P(0): | $h_3=2.25$ | P(1): | l | Solution |
|---|---|---|---|---|
| (0, 4.5) |  | (2.25,2.25) | 2 |  |
| (4.5,0) |  | (2.25,2.25) | 1 |  |
| (-4.5,0) | $\xrightarrow{M}$ | (2.25,2.25) | 0 | (2.25,0) |
| (0, -4.5) |  | (2.25,2.25) | 1 |  |
| (0, 0) |  | (2.25,0) | 0 |  |

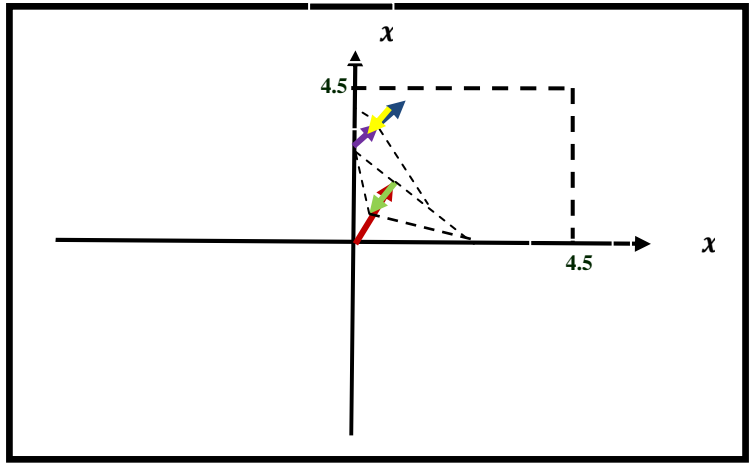

Figure 6 : The process of rotational mutation
and crossover on search space

Table 6: comparison between test problem #2 and other three

| Algorithms | Iteration | Optimal point | Best Solution | Standard deviation |
|---|---|---|---|---|
| **SGM** | 24 | (3,0.5) | | (0,0) |
| **RS** | 1000 | (3,0.5) | | (0,0) |
| **RSW**($x^{initial} =$ (14.0356,14.0356 )) | 450 | ( 3, 0.4865 ) | (3,0.5) | ( 0, 0. 0135 ) |
| **SA** | 120 | (3,0.5) | | (0,0) |

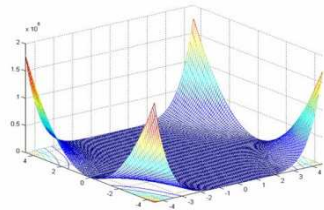

Figure 7 : Beal function's graph

### 5. Schema of SGM:

As compared to other optimization methods, Genetic Algorithm (GA) as an auto-adapted global searching system by simulating biological evolution and the fittest principle in natural environment seeks best solution more effectively [6]. Therefore, GA is a heuristic search technique that mimics the natural evolution process such as initialization, selection, crossover and mutation operations. The selection pressure drives the population toward better solutions while crossover uses genes of selected parents to produce offspring that will form the next generation.

Mutation is used for avoiding of premature convergence and consequently escaping from local optimal. The GAs have been very successful in handling combinatorial optimization problems which are difficult [7].

### 5.1. Initialization

Initially many individual solutions are generated to form an initial population. The population size depends on the nature of the problem; the initial population in SGM is generated according to dimension for example initial population for n- dimensional is $2^n$ according to table 7. Initial population in this method earn through crossing point with equation $2^n$ .In SGM are considered as gen every crossing point.

Table7: The initial population in SGA

| Dimensional | Crossing points |
|---|---|
| 2 | $2^2$ |
| 3 | $2^3$ |
| 4 | $2^4$ |
| ⋮ | ⋮ |
| n | $2^n$ |

### 5.2. Selection

During each successive generation, a proportion of the existing population is selected to create a new generation. Individual solutions are selected through a fitness-based process, where fitter solutions (as measured by a fitness function) are typically more likely to be selected. Certain selection is preferentially considered to be the best solution.
Suppose that algorithm is searching a point x which can make continuous function of *f* to achieve its global point. The necessary and sufficient condition of extreme point is that this point gradient is 0 that is $\nabla f(x) = 0$.

For self-mapping: $\mathbb{R}^n \to \mathbb{R}^n$, we say $x \in \mathbb{R}^n$ is a fixed point of $g$ if $(x) = x$, then we can convert the solution of zero point problems to fixed point ones of function $g(x) = x + \nabla f(x)$.
Suppose that definition domain of $f(x_1, x_2, ..., x_n)$ is $a_1 \leq x_1 \leq a_2$ , $a_3 \leq x_2 \leq a_4$, ..., $a_r \leq x_n \leq a_s$ and divide the domain into many polytopes with two groups of straight lines of $\{x_1 = mh_i\}, \{x_2 = mh_i\}, ..., \{x_n = mh_i\}$ in which m is a non-negative integer and $h_i$ is a positive quantity relating to precision of the problem. As a result, we can code each point of intersection as $x_1 = a_1 + k_1 h_i, x_2 = a_3 + k_2 h_i, ..., x_n = a_r + k_n h_i$ where $k_1, k_2, ..., k_n$ are not negative integers, so $(k_1, k_2, ..., k_n)$ is called the relative coordinates of points. Consequently, by changing $k_1, k_2, ..., k_n$ relative coordinates of each point in search space is determined. Attentive points are pointed above and labeling is done according to equation 6:

$$l(x) = \begin{cases} 0, & g_1(x)-x_1 \geq 0, \ldots, g_n(x)-x_n \geq 0 \\ 1, & g_1(x)-x_1 < 0, g_2(x)-x_2 \geq 0, \ldots, g_n(x)-x_n \geq 0 \\ 2, & g_2(x)-x_2 < 0, g_3(x)-x_3 \geq 0, \ldots, g_n(x)-x_n \geq 0 \\ \vdots & \\ n, & g_n(x)-x_n < 0 \end{cases} \quad (6)$$

The square with all different kinds of integer label is called a completely labeled unit, when $h_i \to 0$ within iteration stages, and vertices of that square approximately converge to one point which is a fixed point.

### 5.3. Mutation Operator for reducing search space

We use mutation operator for reducing search space and we use several time from this operator for minimizing search space and become closer to Mutation Operator perform following:

At first, for each point coded $(k_1, k_2, \ldots, k_n)$, thenGA tries to improve it to reach optimal solution by mutation operator searching all points surrounding it in certain step determined by $h_{i+1}$.

For instance, if $(k_1, k_2, \ldots, k_n)$ is in P (0), initial population, addressing $(x_1 + k_1 h_i, x_2 + k_2 h_i, \ldots, x_n + k_n h_i)$ will be changed as $(x_1 + \alpha_1, x_2 + \alpha_2, \ldots, x_n + \alpha_n), \alpha_1, \alpha_2, \ldots, \alpha_n \in \{0, \pm h_{i+1}\}$. Subsequently, the algorithm saves the best-mutated individual among all possible off-springs. Therefore, this operator produces new population located on intersection of the next grid. Because of this, further Polytopes are specified to evaluate and label.

### 5.4. Rotational MutationOperator

Rotational Mutation Operator combines with crossover for finding global optimal point.in part, we explain Rotational MutationOperator. Rotational Mutation Operator performs following:

Regarding the selected polytope in section 5.3, offspring (vertex) is selected that has best fitness ('S') and mutates offspring 'S' in direction of $\vec{r_0}$ vector with length of mutation α. ( notice: the direction of $\vec{r_0}$ vector must be inside the search space, and the α measurement depends on problem precision).Vector $\vec{r_n}$ is assigned based on search space and mutation size is the same as below:
$\vec{r_0}$=(1,1,…,1),(1,-1,1,…,1),…, or (-1,-1,…,-1)
$\vec{r_n} = \alpha \vec{r_0}$
$\alpha = 0.1n$
n=1, 2,…, 10

When we use mutation, end of mutation vector gives us one point ('p') that if f (p) has best fitness than f(s) and p replaces 'S' point else use rotational Mutation and find point that has better fitness than point S.

## 5.5. Crossover Operator

As we said, Crossover Operator and Rotational Mutation Operator combined together for finding global point. In pervious part, we explained Rotational Mutation Operator.
In this part Crossover Operator is combined with Rotational Mutation Operator that is explained pervious for finding global optimal point.

After using rotational Mutation, crossover operator acts on sides of polytope and produce new descendants as following computations.

In 2-dimensional space, suppose that $P_1 = (x_1^{(1)}, x_2^{(1)})$ and $P_2 = (x_1^{(2)}, x_2^{(2)})$ are two parent vectors. Consider following case as descendant obtained though crossover operator.

$$(\frac{x_1^{(1)} + x_1^{(2)}}{2}, \frac{x_2^{(1)} + x_2^{(2)}}{2})$$

In a 3-dimensional space, supposing that $P_1 = (x_1^{(1)}, x_2^{(1)}, x_3^{(1)})$ and $P_2 = (x_1^{(2)}, x_2^{(2)}, x_3^{(2)})$ are two parent vectors. Consider the following case as descendant obtained though crossover operator.

$$(\frac{x_1^{(1)} + x_1^{(2)}}{2}, \frac{x_2^{(1)} + x_2^{(2)}}{2}, \frac{x_3^{(1)} + x_3^{(2)}}{2})$$

In an n-dimensional space, suppose that $P_1 = (x_1^{(1)}, x_2^{(1)}, \dots,)$ and $P_2 = (x_1^{(2)}, x_2^{(2)}, x_3^{(2)})$ are two parent vectors. Consider the following case as descendant obtained though crossover operator.

$$(\frac{x_1^{(1)} + x_1^{(2)}}{2}, \frac{x_2^{(1)} + x_2^{(2)}}{2}, \frac{x_3^{(1)} + x_3^{(2)}}{2}, \dots, \frac{x_n^{(1)} + x_n^{(2)}}{2})$$

## 6. Schema Analysis for SGM

In SGM, at first, search space should minimize by mutation operator according to Equation 4 to be closer to solution. Then, in this search space, there is found an optimal point with rotational mutation and crossover so that we can present a schema that have two part one of them is for reducing search space by mutation operator and another is for finding global optimal point by Crossover Operator and Rotational Mutation Operator. Our flowchart is based on two part. One part focused on reducing search space and second focused on finding global point.

Schema in Figure8, at first part, makes grids in given scope and encode each intersection point by integer number while it starts from the lowest point of the domain. After calculating fitness of each point, it generates the best offspring and computes integer label of the last population for every Polytope. When it found completely labeled polytope, we subdivide it in order to seek the solution closely.

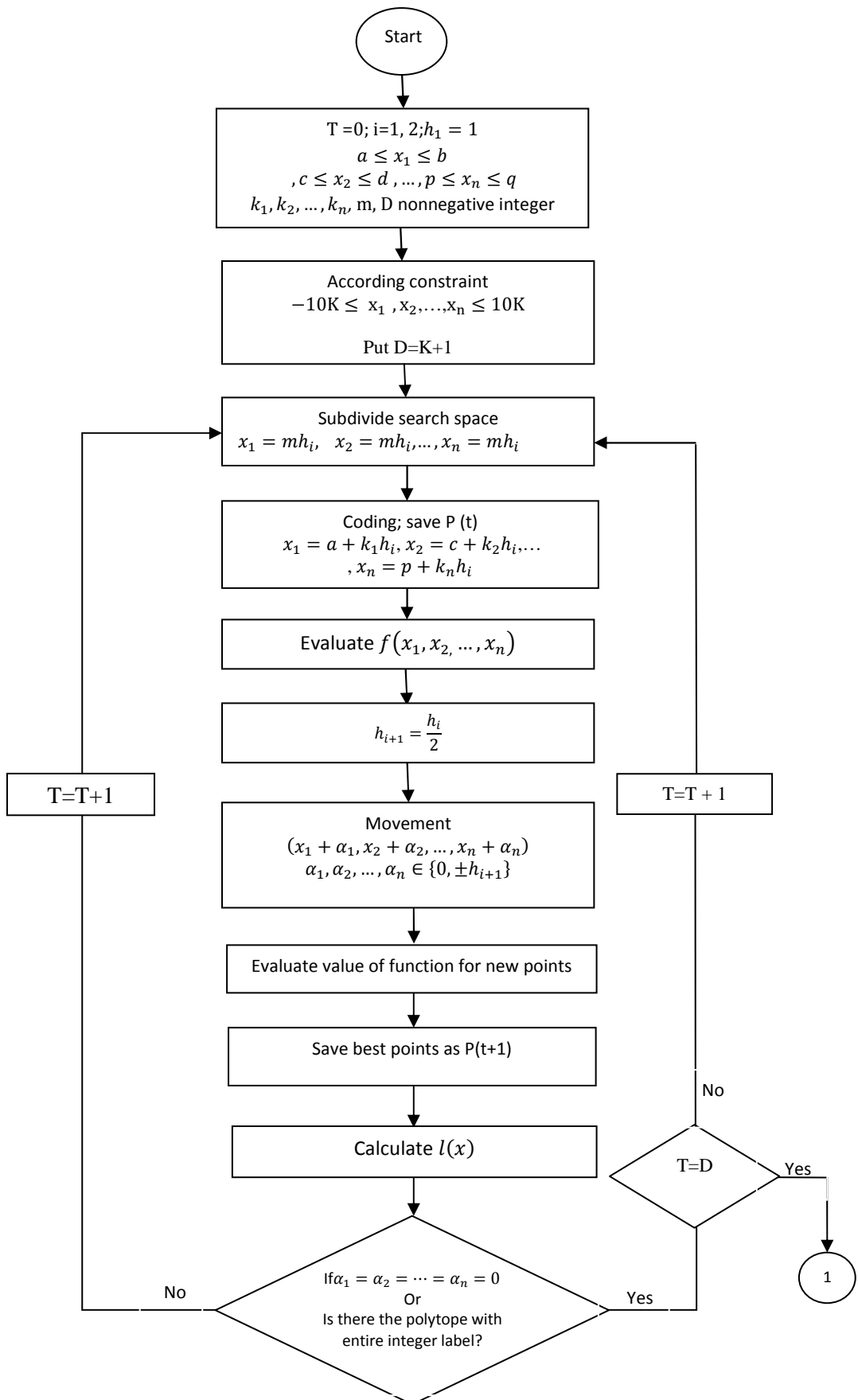

**Figure 8: schema for reducing search space**

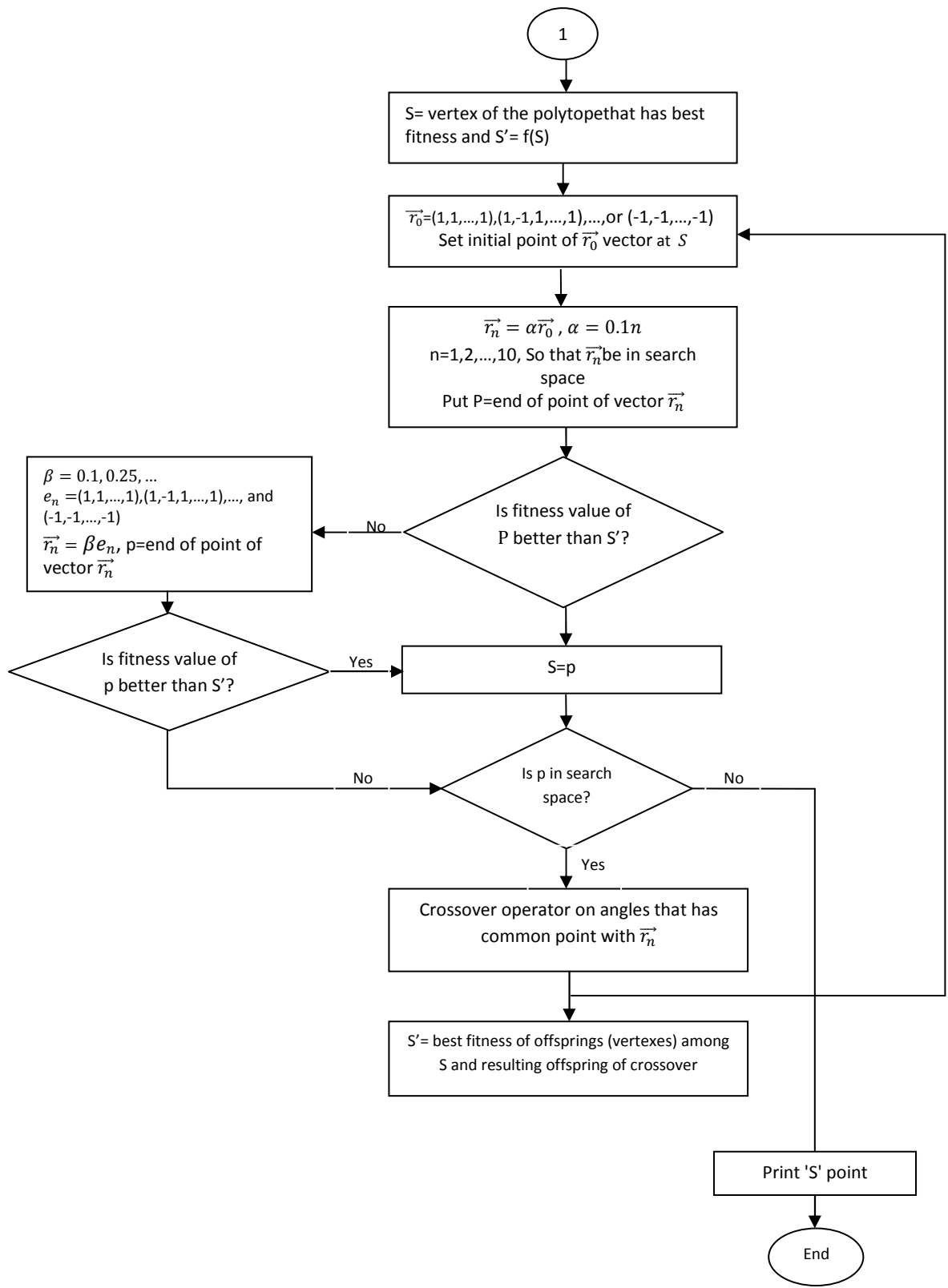

**Figure 9: schema for finding optimal point by rotational mutation**

At Second part, Figure9,SGM starts from the completely labeled polytope-the result of the first schema-. It selects anoffspring (vertex) which has best fitness value that is called 'S'.Then, it sets initial point of $\vec{r_0}$ vector at 'S' and mutates offspring 'S' in direction of $\vec{r_0}$ vector with length of mutation α.(Notice: the direction of $\vec{r_0}$ vector must be inside the search space, also the α measurment depends on problem precision). This mutated offspring is called $p$.

If fitness value of offspringp isbetter than fitness value of offspring S, we can use crossover operator for adjacent sides of offspring $p$. Otherwise, by using rotational mutation by $\vec{e_n}$vector, we would search an Offspring-with better fitness value in comparison witch's'. Then, we make a crossover. After that, we select an offspring with the better fitness value – between mutated offspring and crossover offspring's. We mutate and make a crossover on it again. We repeat this action while the mutated offspring doesn't get out of search space. Eventually, we would select the last new produced offspring- inside the search space- as the global optimization point which would be the fixed point of our question.

## 7. Evaluation

In this section, Definition of De Jong's Functions, Numerical results of De Jongs' functions bySGMand thecompression of SGMwith the other methods (DE, PGA, Grefensstette and Eshelman) for De Jong Functions are presented.

Table 8: De Jong's Functions

| Function Number | Function | Limits | Dim. | Initial Population |
|---|---|---|---|---|
| F1 | $\sum_{i=1}^{3} x_i^2$ | $-5.12 \leq x_i \leq 5.12$ | 3 | 8 |
| F2 | $100.(x_1^2 - x_2) + (1 - x_1)^2$ | $-2.048 \leq x_i \leq 2.048$ | 2 | 4 |
| F3 | $30. + \sum_{i=1}^{5} \lfloor x_j \rfloor$ | $-65.536 \leq x_i \leq 65.536$ | 5 | 32 |
| F4 | $\sum_{i=1}^{30} (ix_i^4 . + Gauss\ (0,1))$ | $-1.28 \leq x_i \leq 1.28$ | 30 | |
| F5 | $\dfrac{1}{0.002 + \sum_{i=0}^{24} \dfrac{1}{i+\sum_{j=0}^{1}(x_j - a_{ij})^6}}$ | $-65.536 \leq x_i \leq 65.536$ | 2 | 4 |

### 7.1De Jong's Functions

In this section, Definition of De Jong's Functions and their graphs are shown.

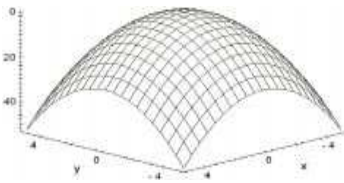
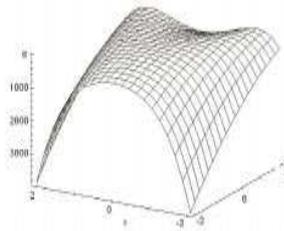
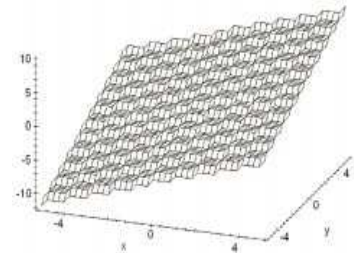

**Figure 10: F1 Sphere**  **Figure 11: F2 Rosenbrock's**  **Figure 12: F3 Step Function.**

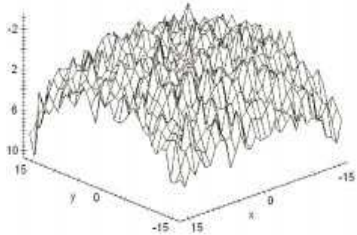
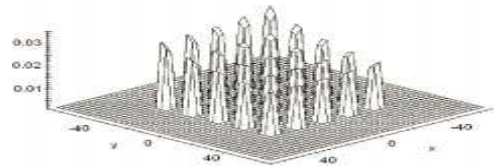

**Figure 13: F4 Stochastic Function.**  **Figure 14: F5 Foxholes Function.**

### 7.2 Numerical Results

In this section, we implemented SGM on the five problems of De Jong (F1 to F5) [De Jong, 1975] and results are shown in Table ……

In this experimental are examined: mutation size (MS), mutation rate (MR), rotational mutation size (RMS), times of repetition of rotational mutation (TRM), times of repetition of crossover (TC) andtimes of repetition of first schema (TF).
Complete label selection procedure and population size depend on dimensional space according to Table 8. As Table 9 shows, for all functions, each experiment consists of running the genetic algorithm for 1000 trials (function evaluations).Results were saved for the best performance (BP), BP is the smallest value of the objective function obtained over all function evaluations. Atlast, standard deviation (SD) is calculated and measured with final answer of De Jong function.

The average number of generations presented in Tables 8, 9 provides the minimum function values inthree-digit accuracy. We have compared the performance of the PGA to that of other some well-known genetic algorithms: DE, PGA, Grefensstette and Eshelman [2]. SGA, Grefensstette and Eshelman algorithms were run 50 times; the DE algorithm was run 1000 times and PGA 100 times for each function to achieve average results[2].
When the results produced by the SGA for all functions were evaluated together, it was Observed that the best of the average of number of generations for De Jong'sfunctions.
It is seen in Table 9 that the convergence speed of the SGA achieves the optimal point with more decision by smaller generation and SGA has simple operation than other methods.

As Table 10 the most significant improvement is with F5, proportion of the number of generations (PNG) about $\frac{1200}{19} \cong 64$ time's smaller average of number of generations than the DE algorithm.

Table9: Best Values of Best Performance

| Step | Algorithm | De Jong's Function | TF | MR | RMS | TRM | TC | Best Point | BP | SD |
|------|-----------|--------------------|-----|-----|-----|-----|-----|------------|-----|-----|
| Result | RSLMGA | F1 | 2 | 0.5 | 0.1 | 15 | 3 | (0,0,0) | 0 | 0 |
| Result | RSLMGA | F2 | 2 | 0.5 | 0.1 | 16 | 11 | (1,1) | 0 | 0 |
| Result | RSLMGA | F3 | 2 | 0.5 | 0.1 | 25 | 5 | (-5.12,-5.12,-5.12, -5.12, -5.12) | 0 | 0 |
| Result | RSLMGA | F4 | 2 | 0.5 | 0.1 | 75 | 30 | (0,0,…,0) 30 | Depend on $\eta$ | 0 |
| Result | RSLMGA | F5 | 8 | 0.5 | 0.1 | 9 | 2 | (-32,-32) | 0 | 0 |

Table 8: Numerical result of De Jongs' function by RSLMGA

| Algorithms | F1 | F2 | F3 | F4 | F5 |
|------------|------|-------|------|------|------|
| PGA($\lambda = 4$) | 1170 | 1235 | 3481 | 3194 | 1256 |
| PGA($\lambda = 8$) | 1526 | 1671 | 3634 | 5243 | 2076 |
| Grefensstette | 2210 | 14229 | 2259 | 3070 | 4334 |
| Eshelman | 1538 | 9477 | 1740 | 4137 | 3004 |
| DE(F: RandomValues) | 260 | 670 | 125 | 2300 | 1200 |
| RSLMGA | 20 | 29 | 32 | 107 | 19 |
| PNG | 13 | 24 | 4 | 22 | 64 |

## 8. Conclusion

In this paper, we proposed a new method for optimization function is based on subdividing that is implemented by GA. SGM use several time from mutation operator for minimizing search space and became closer to solution then focused on global optimal point and find final point by combination of rational mutation and crossover.

The performance of the SGM has been compared to that of some other well-known GA methods such as Grefensstette, Random Value, and PNG. Simulations of results show that, it was observed that SGM achieve the optimal point with more decision by smaller generation. Therefore, SGM seems to be a promising approach for engineering optimization problems.